%% file: main.tex
\documentclass{article}

\usepackage{amsmath}
\usepackage{amssymb}
\usepackage{mathtools}
\usepackage{amsthm}
\input{math_commands.tex}

\usepackage{microtype}
\usepackage{graphicx}
\usepackage{subfigure}
\usepackage{booktabs} % for professional tables

\usepackage{hyperref}

\usepackage{colortbl}

\usepackage[accepted]{icml2025}

\usepackage{amsmath}
\usepackage{amssymb}
\usepackage{mathtools}
\usepackage{amsthm}

\usepackage[capitalize,noabbrev]{cleveref}

\theoremstyle{plain}

\theoremstyle{definition}

\theoremstyle{remark}

\usepackage{paralist}

\usepackage[textsize=tiny]{todonotes}

\begin{document}

\twocolumn[
\icmltitle{Exploring Representation-Aligned Latent Space for Better Generation}



\icmlsetsymbol{equal}{*}

% Wanghan Xu, Xiaoyu Yue, ZiDong Wang, Yao Teng, zhangwenlong, Xihui Liu, Luping Zhou, Wanli Ouyang, LEI BAI

\begin{icmlauthorlist}
\icmlauthor{Wanghan Xu}{a,b}
\icmlauthor{Xiaoyu Yue}{b,c}
\icmlauthor{Zidong Wang}{b,d}
\icmlauthor{Yao Teng}{e}
\icmlauthor{Wenlong Zhang}{b}
\icmlauthor{Xihui Liu}{e}
\icmlauthor{Luping Zhou}{c}
\icmlauthor{Wanli Ouyang}{b}
\icmlauthor{Lei Bai}{b}
\end{icmlauthorlist}

\icmlaffiliation{a}{Shanghai Jiao Tong University}
\icmlaffiliation{b}{Shanghai AI Laboratory}
\icmlaffiliation{c}{University of Sydney}
\icmlaffiliation{d}{Chinese University of Hong Kong}
\icmlaffiliation{e}{University of Hong Kong}

% \icmlcorrespondingauthor{}{}

\icmlkeywords{Generation}

\vskip 0.3in
]



\printAffiliationsAndNotice{} % otherwise use the standard text.

\input{sections/abstract}

\input{sections/introduction}
\input{sections/related_work}
\input{sections/method}
\input{sections/experiments}

\input{sections/conclusion}

\newpage

\nocite{langley00}

\input{sections/impact_statements}

\bibliography{main}
\bibliographystyle{icml2025}

\input{sections/supply}

\end{document}

%% file: math_commands.tex
\usepackage{amsmath,amsfonts,bm}

\def\eqref#1{equation~\ref{#1}}
\def\1{\bm{1}}

\def\vx{{\bm{x}}}

\def\vz{{\bm{z}}}

\def\mI{{\bm{I}}}

\DeclareMathAlphabet{\mathsfit}{\encodingdefault}{\sfdefault}{m}{sl}
\SetMathAlphabet{\mathsfit}{bold}{\encodingdefault}{\sfdefault}{bx}{n}

\def\gL{{\mathcal{L}}}

\def\sR{{\mathbb{R}}}

%% file: sections/abstract.tex
\begin{abstract}

Generative models serve as powerful tools for modeling the real world, with mainstream diffusion models, particularly those based on the latent diffusion model paradigm, achieving remarkable progress across various tasks, such as image and video synthesis. Latent diffusion models are typically trained using Variational Autoencoders (VAEs), interacting with VAE latents rather than the real samples. While this generative paradigm speeds up training and inference, the quality of the generated outputs is limited by the latents' quality. Traditional VAE latents are often seen as spatial compression in pixel space and lack explicit semantic representations, which are essential for modeling the real world.
In this paper, we introduce \textbf{ReaLS} (Representation-Aligned Latent Space), which integrates semantic priors to improve generation performance. Extensive experiments show that fundamental DiT and SiT trained on ReaLS can achieve a \textbf{15\%} improvement in FID metric. Furthermore, the enhanced semantic latent space enables more perceptual downstream tasks, such as segmentation and depth estimation. Code and model checkpoints are available at \textcolor{pink}{https://github.com/black-yt/ReaLS} .

\end{abstract}

%% file: sections/introduction.tex
\section{Introduction}

% generation and diffusion models
The objective of generative models is to accurately capture and model the distribution of the real world, enabling the creation of outputs that are not only visually compelling but also semantically coherent. Existing diffusion-based generative models~\cite{dit, chang2022maskgit, rombach2022ldm} typically sample from a random distribution, e.g., a Gaussian distribution, and then iteratively refine the samples to approximate the distribution of the real world. These models achieve remarkably successful results in fields such as image, audio, and video generation~\cite{bar2023multidiffusion, huang2023make, ho2022video}.

\begin{figure}[t]
    \centering
    \includegraphics[width=0.90\linewidth]{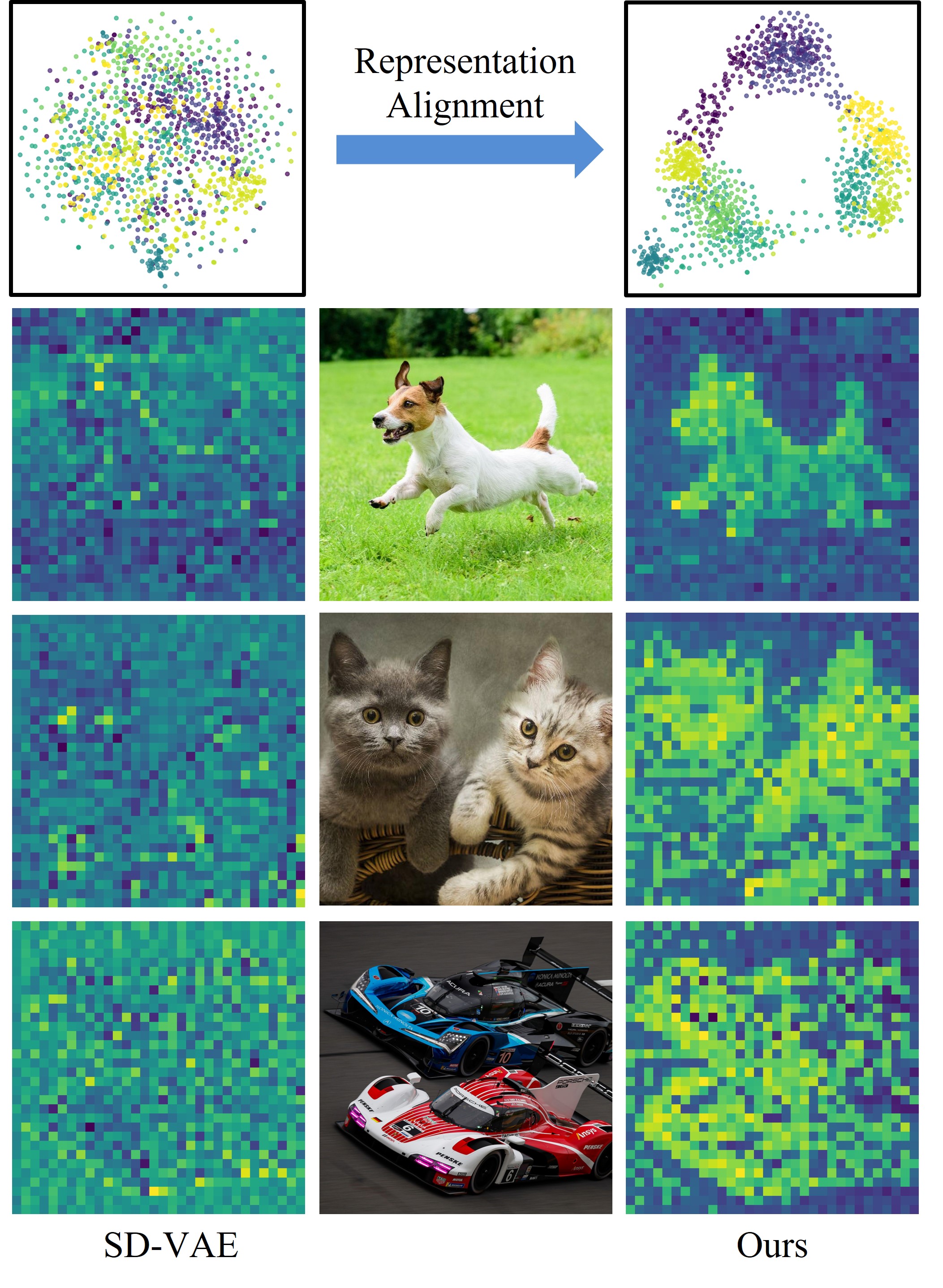}
    \vspace{-0.6cm}
    \caption{\textbf{Representation-Aligned Latent Space (ReaLS) preserves more image semantics.} a) t-SNE visualization of our latent space reveals a clear clustering, with samples from the same category closer to each other. b) Attention map of our latents shows a significant improvement in the semantic relevance among patches.}
    \label{fig:cluster}
    \vspace{-0.7cm}
\end{figure}

Latent diffusion models (LDMs)~\cite{rombach2022ldm}, as a typical type of generative model, commonly utilize Variational Autoencoders (VAEs)~\cite{doersch2016tutorial} to enhance training and inference efficiency. VAEs first encode real samples into a latent space with spatial compression, where diffusion is performed to fit the latent distribution. However, the capability of the VAE's modeling of the real world limits the quality of the final samples generated by the LDM. Therefore, \textbf{\textit{developing a more effective latent space for diffusion models is essential, yet remains underexplored}}.

Traditional VAEs are optimized to compress images into more compact latent representations, prioritizing local textures at the expense of global image context. This local encoding property results in the VAE latent lacking rich semantic information about the images, which is crucial for perceiving the real world~\cite{yu2024representation}. 

To specifically illustrate the limitations of traditional latent space, we present t-SNE and attention map visualizations of SD-VAE~\cite{rombach2022high}, a widely used VAE in LDMs. Figure~\ref{fig:cluster} reveals two key observations: a) t-SNE visualization indicates that it struggles to represent the characteristics of different categories within the latent space; b) the attention maps of the latents show that it fails to capture the relationships between different parts of the same instance. These observations highlight the lack of semantic representation in SD-VAE, which hinders LDM learning. Consequently, although the generated outputs may appear visually plausible, they often fall short of achieving semantic congruence with the intended descriptions or tasks. 

In this work, we construct a semantically rich latent space through a new VAE training strategy, which not only compresses the original image but also preserves the inherent relationships within the data. Unlike traditional VAEs that apply KL constraints~\cite{doersch2016tutorial} solely in the latent space, we align the VAE's latents with features from DINOv2~\cite{oquab2023dinov2}, explicitly injecting semantic representations of images into the latent space. During training, we found that the quality of images generated by LDMs is closely related to the balance between the KL divergence constraint and alignment constraint in the latent space. This is because the KL constraint and the alignment constraint provide guidance for unity and differentiation, respectively. The former focuses on the overall consistency of the latents, driving them toward a standard normal distribution, while the latter considers the semantic differences between samples. When these two constraints are balanced, the latent space approximates a standard normal distribution while retaining the semantic features, as illustrated in Figure~\ref{fig:cluster}.

Extensive experiments demonstrate that existing generative models, such as DiT~\cite{dit} and SiT~\cite{sit}, benefit from Representation-Aligned Latent Space (\textbf{ReaLS}) without requiring modifications. It achieves a notable 15\% improvement in FID performance for image generation tasks. Additionally, the richer semantic representations in the latents enable more downstream perceptual tasks, such as image segmentation~\cite{minaee2021image} and depth estimation~\cite{ming2021deep}.

We summarize the contributions of this paper as follows:
\vspace{-0.15cm}
\begin{compactitem}
  \item We propose a novel representation-alignment VAE, which provides a better latent space for latent generative models with semantic priors.
  \item Representation-Aligned Latent Space (\textbf{ReaLS}) can significantly improve generation performance of existing LDMs without requiring any changes to them.
  \item The semantically rich latent space enables downstream perceptual tasks like segmentation and depth detection.
\end{compactitem}

%% file: sections/related_work.tex
\section{Related Work}
\subsection{Variational Autoencoders in LDM Paradigm} 
% \paragraph{Variational Autoencoders for Visual Generation.} 
Stable Diffusion~\cite{rombach2022ldm} introduced the latent diffusion model paradigm, employing a variational autoencoder (VAE)~\cite{kingma2013vae} (SD-VAE), the most widely used VAE, to encode visual signals from image space into latent space and decode these latent tokens back into images. This approach has facilitated the training and scaling of diffusion models, establishing itself as the dominant choice for visual generation. The quality of the VAE sets the upper limit for generative models, prompting significant efforts to enhance VAEs. SDXL~\cite{podell2023sdxl} retains the SD-VAE architecture while adopting advanced training strategies to improve local and high-frequency details. LiteVAE~\cite{sadat2024litevae} utilizes the 2D discrete wavelet transform to boost scalability and computational efficiency without compromising output quality. SD3~\cite{esser2024sd3} and Emu~\cite{dai2023emu} expand the latent channels of VAEs to achieve better reconstruction and minimize information loss. DC-AE~\cite{chen2024dcae} and LTX-Video~\cite{hacohen2024ltxvideo} increase the compression ratio while maintaining satisfactory reconstruction quality. 

These VAEs often focus on improving image compression and reconstruction. However, we found that better reconstruction does not necessarily lead to better generation (discussed in detail in Section~\ref{discussion}). This paper explores enhancing the generation quality of LDMs by injecting semantic representation priors into the latent space, providing new insights for improving VAE training.

% \vspace{-0.2cm}

\subsection{Diffusion Generation and Perception} 

Beyond image generation, diffusion models have been increasingly applied to a variety of downstream perceptual tasks. VPD~\cite{zhao2023vpd} leverages the semantic information embedded in pre-trained text-to-image diffusion models, utilizing additional specific adapters for enhanced visual perception tasks. Marigold~\cite{ke2024marigold} repurposes a pre-trained Stable Diffusion model into a monocular depth estimator through an efficient tuning strategy. JointNet~\cite{zhang2023jointnet} and UniCon~\cite{li2024unicon} employ a symmetric architecture to facilitate the generation of both images and depth, incorporating advanced conditioning methods to enable versatile capabilities across diverse scenarios. SDP~\cite{ravishankar2024sdp} utilizes a pre-trained DiT-MoE model on ImageNet, exploring the advantages of fine-tuning and test-time computation for perceptual tasks. 

The models mentioned above do not seek to unify generation and perception within the latent space. In this work, LDM trained on ReaLS is inherently rich in semantics and enables training-free execution of downstream perceptual tasks, including segmentation and depth estimation.

%% file: sections/method.tex
\section{Method}

\paragraph{Overview.} As a leading paradigm in generative modeling, the latent diffusion model (LDM)~\cite{rombach2022ldm} operates in latent space. During training, a visual encoder first reduces the image from the original pixel space to the latent space. Diffusion model is then trained in this latent space through processes of adding noise and denoising. In the generation phase, LDM iteratively denoises the sampled latent noise into a clean latent representation, which is then converted into an image using a corresponding decoder. Traditional latent spaces primarily serve as spatial compressors and often lack the semantic information which is crucial for generation tasks. This work enhances the latent space by aligning semantic representations within a VAE, resulting in a more robust semantic structure that not only improves the quality of diffusion-generated images but also facilitates downstream tasks such as segmentation and depth detection.

\begin{figure*}
    \centering
    \includegraphics[width=0.99\linewidth]{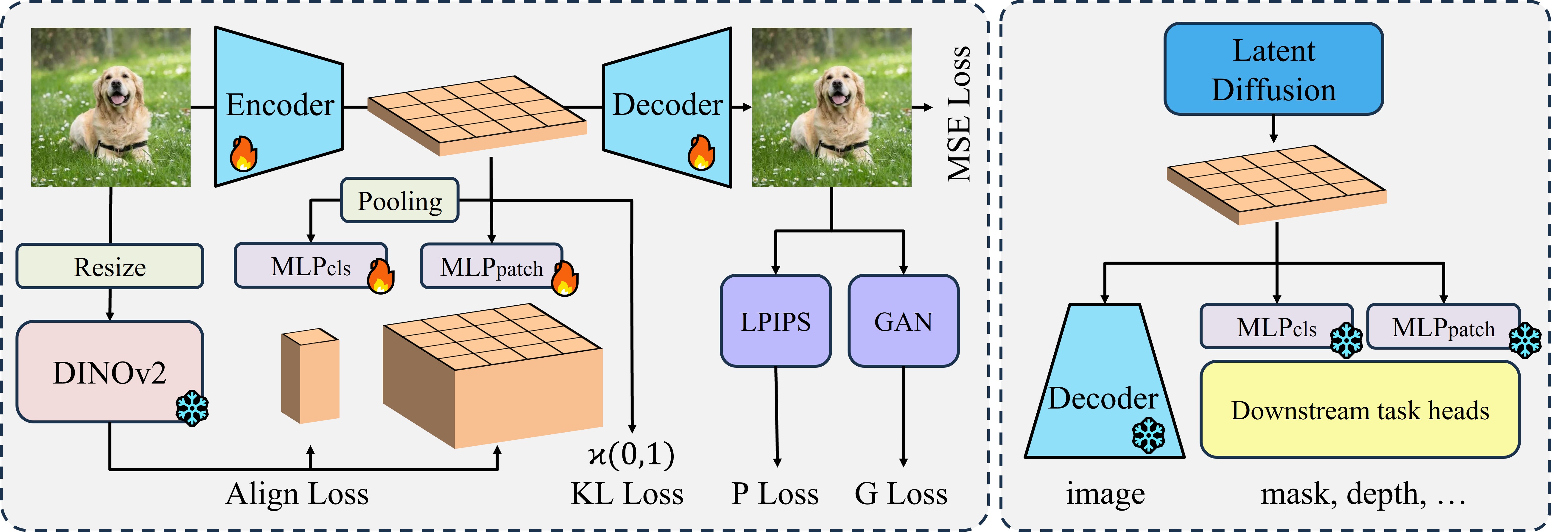}
    \caption{\textbf{The training and inference pipeline of ReaLS.} During VAE training, the latents of the VAE are aligned with the features of DINOv2 using an alignment network implemented via MLP. After the VAE training concludes, latent diffusion model training is performed in this latent space. In the inference phase, the latents generated by the diffusion model are converted into corresponding generated images through the VAE decoder. At the same time, the alignment network extracts semantic features, which are provided to the corresponding downstream task heads, enabling training-free tasks such as segmentation and depth estimation.}
    \label{fig:framework}
    \vspace{-0.5cm}
\end{figure*}

\subsection{Preliminary}
\label{Sec:preliminary}
\paragraph{Variational Auto-Encoder.} 

Variational Autoencoders (VAE)~\cite{doersch2016tutorial} are a type of generative model that encodes images from pixel space to latent space by learning image reconstruction.
Let  $\vx \in \sR^{3 \times H \times W}$ represent an RGB image, where \(H\) and \(W\) denote its height and width, respectively. A VAE typically consists of two main components: an encoder and a decoder. The role of the encoder is to map the input data \(\vx\) to a latent space \(\vz \in \sR^{\frac{H}{p} \times \frac{W}{p} \times D}\) that follows a Gaussian distribution, where \(p\) represents patch size. This mapping is mathematically represented as:
% \begin{equation}
$q_\phi(\vz|\vx) = \mathcal{N}(\vz; \mu_\phi(\vx), \sigma^2_\phi(\vx) \mI) , $
% \end{equation}
where \(\mu_\phi(x)\) and \(\sigma^2_\phi(x)\) are computed by a neural network parameterized by \(\phi\). 
% During this process, it is essential for \(\vz\) to approximate a standard normal distribution, as this ensures the generative model's ability to generalize, meaning that noise sampled from the normal distribution can be decoded into high-quality images.
During this process, \(\vz\) approximately satisfies a standard normal distribution, so a noise sampled from the normal distribution can be decoded into a high-quality image.
Therefore, a KL divergence loss constraint is added, as shown in the following formula:
% \begin{equation}
$\mathcal{L}_{KL} = D_{KL}(q_\phi(\vz|\vx) || p(\vz)).$
% \end{equation}

After the encoding process, the decoder reconstructs the original data from the latent representation. It models the data distribution based on the latents to generate new samples:%, as follows:
% \begin{equation}
$p_\theta(\vx|\vz) = \mathcal{N}(\vx; \mu_\theta(\vz), \sigma^2_\theta(\vz) \mI) , $
% \end{equation}
where \(\mu_\theta(\vz)\) and \(\sigma_\theta(\vz)\) is computed by a neural network parameterized by \(\theta\). This collaborative structure between the encoder and decoder makes VAE a powerful tool in generative modeling.

\paragraph{Latent Diffusion Model.} 

Latent Diffusion Models (LDM) are a type of diffusion model trained in the latent space.
During training, LDM learns to predict the noise in the input latents that have been perturbed by various levels of Gaussian noise. During inference, starting from pure Gaussian noise, the LDM progressively removes the predicted noise and ultimately obtains a clean latent.
Since LDMs generate data in latent space, a well-structured latent space that incorporates both low-level pixel information and high-level semantic information is crucial for high-quality image generation.
In this paper, we demonstrate that by aligning with semantics, we can construct a more structured latent space, effectively enhancing the quality of the generated outputs.

% first adds noise to the latent representation \( \vz\) and then learns the denoising process through a network.
% Taking noise prediction as an example, LDM first adds noise to the latent representation \( \vz\) and then learns the denoising process through a network.
% This can be expressed with the following formulas:
% \begin{equation}
% z_{t-1} = z_t - \epsilon_\theta (z_t, t)
% \end{equation}

\subsection{Representation Alignment}

Traditional VAEs compress images into latent space through reconstruction tasks, resulting in a latent space that serves merely as a compressed representation of pixel data and lacks crucial semantic information. We enhance VAE training by incorporating semantic representation alignment, enriching the latent space with semantic content, which facilitates diffusion generation within this space.

Specifically, we use DINOv2~\cite{oquab2023dinov2} as the image semantic representation extractor. For an input image \( \vx \in \sR^{3 \times H \times W}\), DINOv2 outputs two types of features: a)~the image patch feature, denoted as \(\mathcal{F}_\text{p} \in \sR^{\frac{H}{p'} \times \frac{W}{p'} \times D'}\) where \(p'\) is the patch size of DINOv2; b)~the global image feature, denoted as \( \mathcal{F}_{\text{cls}} \in \mathbb{R}^{D'}\).
To align with \(\mathcal{F}_\text{p}\), we ensure that the patches obtained from DINOv2 have a one-to-one correspondence with VAE latents. Formally, we resize the image to \((H', W')\) before feeding it into DINOv2, where \((\frac{H'}{p'}, \frac{W'}{p'}) = (\frac{H}{p}, \frac{W}{p})\).
To align with \(\mathcal F_{\text{cls}}\) that reflects the global semantics of the image, such as object categories, we average the VAE latents across the spatial dimensions to gather the global information of the image.

Subsequently, through two align networks implemented with Multilayer Perceptron (MLP), we map the latents from dimension $D$ to the DINOv2 feature dimension $D’$:

\vspace{-0.5cm}
\begin{equation}
\left\{\begin{matrix}
\mathcal F_{\text{vae},ij}=\text{MLP}_{\text{patch}}(\vz_{ij})    \\
\mathcal F_{\text{vae},\text{cls}}=\text{MLP}_{\text{cls}}(\text{AP}(\vz))
\end{matrix}\right. ,
\vz = \mu_\phi(\vx)+\sigma_\phi(\vx)\epsilon,
\label{equ:alignnet}
\end{equation}
where \(\mu_\phi(\vx)\) and \(\sigma_\phi(\vx)\) are the mean and variance estimated by the VAE encoder, \(\vz\) is obtained by the reparameterization, \(\epsilon\) is a random noise, $\text{AP}(\cdot)$ denotes average pooling.

For the alignment loss, we use a combination of cosine similarity loss and smooth mean squared error (MSE) loss:
\begin{equation}
\gL_{\text{align}} = \lambda_1 \gL_{\text{cos}}(\mathcal F_{\text{vae}}, \mathcal F_{\text{dino}}) + \lambda_2 \gL_{\text{smMSE}}(\mathcal F_{\text{vae}}, \mathcal F_{\text{dino}}) .
\end{equation}
In actual experiments, we set $\lambda_1=0.9$ and $\lambda_2=0.1$.

\subsection{Optimization Objectives}
\label{sec:loss}
The training loss of the VAE can be divided into two parts. The first part is on the pixel space, which ensures that the reconstructed image is consistent with the original image. To improve the quality of the reconstructed image and prevent blurriness, the reconstruction loss also incorporates adversarial loss~\cite{creswell2018generative} and perceptual loss (LPIPS)~\cite{rad2019srobb}, as shown below:
% \vspace{-0.2cm}
\begin{equation}
\gL_{\text{pixel}} = \gL_{\text{MSE}} + \lambda_g \gL_{\text{GAN}} + \lambda_p \gL_{\text{perceptual}} .
\end{equation}
The second part of the loss is on the latent space. In traditional VAEs, a KL divergence loss is typically applied to the latents to ensure that \(\vz\) approximates a standard normal distribution \(\mathcal{N}(\mathbf{0},\mI)\). The KL loss enhances the cohesion of the latent space, allowing \(\vz\) obtained from different images to share a same space, which facilitates the diffusion model to sample and denoise from a normal distribution. Additionally, we introduce semantic constraints on the latent space through our alignment network, imparting semantic priors to $\vz$. As shown in Figure~\ref{fig:cluster}, our VAE exhibits a clear clustering in the latent space, despite not using image class labels during training. In summary, the loss on the latent space can be expressed in the following form:
% \vspace{-0.2cm}
\begin{equation}
\gL_{\text{latent}} = \lambda_{k} \gL_{\text{KL}} + \lambda_{a} \gL_{\text{align}} .
\label{equ:latent loss}
\end{equation}
\(\gL_{\text{latent}}\) guides the construction of an improved latent space from two dimensions. The KL loss constrains the overall integrity of the latent space, independent of individual samples. In contrast, \(\gL_{\text{align}}\) applies to each sample, enabling different semantic samples to exhibit diversity while making similar semantic samples have similar representations. Further analysis of these two losses on the latent space and the final generated quality will be discussed in the Section~\ref{sec:kl analysis}.
Finally, the total training loss is as follows:
% \vspace{-0.3cm}
\begin{equation}
\gL = \gL_{\text{pixel}} + \gL_{\text{latent}} .
\end{equation}

\subsection{Generation with Downstream Tasks.}

After completing the VAE model training, we proceed to train the diffusion model in the latent space. To highlight the improvement in generation quality from the semantically aligned latent space, we do not modify the architecture or training process of the diffusion models.

The semantically aligned latent space provides enhanced semantic priors for generation, improving the quality of the model outputs. Additionally, since the diffusion model is trained in this semantically rich latent space, the generated latents are equipped for various perceptual tasks, such as semantic segmentation and depth estimation.
Specifically, the latents produced by the diffusion model can be mapped to features in the DINOv2 dimension via the alignment network used during VAE training. With the corresponding segmentation and depth estimation heads, we can directly obtain the segmentation masks and depth information for the generated images, as shown in Figure~\ref{fig:visual_2}. This not only demonstrates that the latent space captures richer semantic features through the alignment loss, but also expands the applicability of the generative model to downstream tasks.

%% file: sections/experiments.tex
\section{Experiment}

Through extensive experiments, we aim to validate the following questions:

\begin{compactitem}
  \item Does the latent space of our VAE possess richer semantics and a more structured arrangement compared to traditional VAE spaces?
  \item Is the representation aligned latent space beneficial for generation?
  \item Can the diffusion model trained on the ReaLS effectively perform downstream tasks?
\end{compactitem}

\subsection{Experimental Setup}

\paragraph{Implementation Details.} Our model training is divided into two phases. The first phase is VAE training, followed by latent diffusion training in the second phase. In the first phase, we load SD-VAE~\cite{rombach2022high} which is widely used in LDMs as the pre-trained parameters and then train it on ImageNet~\cite{deng2009imagenet}. We employ DINOv2-large-reg~\cite{oquab2023dinov2} as our semantic extraction model, and utilize a two-layer MLP with GeLU activation functions as the alignment network. In the second phase, we strictly adhere to the training methods of DiT and SiT to ensure a fair comparison. \cref{hyperpara} presents the hyperparameters used in both training phases.

\begin{table}[ht]
\vskip -0.15in
\centering
\setlength{\tabcolsep}{5pt}
\small % \scriptsize % \scriptsize %\footnotesize % \tiny
\caption{\textbf{Training Hyperparameter.}}
% \begin{center}
% \begin{small}
% \begin{sc}
% \resizebox{7cm}{!}{
\begin{tabular}{cccccc}
\toprule
Optimizer & lr   & Schedule & min\_lr & Batch Size & Epoch \\ \midrule
\multicolumn{6}{c}{VAE Training}                           \\ 
AdamW     & 5e-5 & Cosine      & 0.0     & 64         & 10    \\ \midrule
\multicolumn{6}{c}{DiT/SiT Training}                       \\ 
AdamW     & 1e-4 & -        & -       & 256        & -     \\ 
\bottomrule
\end{tabular}
% }
% \end{sc}
% \end{small}
% \end{center}
\label{hyperpara}
\vskip -0.1in
\end{table}

\paragraph{Evaluation.} To validate the semantic capability of the VAE, we designed a new metric based on the latent similarity after different augmentations, denoted as semantic consistency (SC). Its calculation is shown in~\cref{alg:sc}.

\begin{algorithm}[h]
\caption{Semantic Consistency (SC)}
\label{alg:sc}
\begin{algorithmic}
\STATE $x_1 \gets \text{RandomAugmentation}(x)$
\STATE $x_2 \gets \text{RandomAugmentation}(x)$
\STATE $z_1 \gets \text{VAE.encode}(x_1)$
\STATE $z_2 \gets \text{VAE.encode}(x_2)$
\STATE $\text{SC} \gets \text{CosineSimilarity}(z_1, z_2)$
\end{algorithmic}
\end{algorithm}

For traditional VAEs, since they merely compress images, the differences in pixel values after applying two different data augmentations lead to different latent representations for the same image, resulting in a lower SC value. In contrast, our VAE incorporates semantic information, so although the images undergo different data augmentations, their semantics do not change significantly. Therefore, $z_1$ and $z_2$ are closer together in the latent space.

For the generative model, we evaluate its quality with Fréchet Inception Distance (FID)~\cite{fid}, sFID, Inception Score (IS), precision (Pre.), and recall (Rec.), with all metrics assessed on the generated 50,000 samples.

\paragraph{Sampler.} We use the SDE Euler sampler with 250 steps for SiT and set the last step size to 0.04.

\paragraph{Baseline.} We use DiT~\cite{dit} and SiT~\cite{sit} as baseline models. Specifically, we trained four models, that is DiT-B/2, SiT-B/2, SiT-L/2, and SiT-XL/2, on our VAE. These models did not undergo any modifications to their network architecture or hyperparameters.

\subsection{Representation Aligned Latent Space}

We provide evidence from three experiments that our VAE latent space contains richer semantic information. 

% 第一，vae latents的聚类可视化
First, we randomly select 10 categories from ImageNet, with 128 images per category, and obtain their latent representations through VAE encoding, which are then reduced in dimensionality using t-SNE. The visualization in Figure~\ref{fig:cluster} clearly shows that, compared to traditional VAEs, our VAE exhibits significant clustering of categories in the latent representations. This indicates that our latent space has better structural properties, with images from the same category being closer together in the space.

% 第二，attention map可视化
Second, we visualize the attention map between one token $z_{ij}$ and all tokens from the VAE latents. The visualization results in Figure~\ref{fig:cluster} show that Our VAE preserves more semantic information in latent space, with tokens from the same object exhibiting higher similarity.

% 第三，用刚才设计的“语义一致性”指标做个定量分析
Third, we conduct a quantitative analysis of the semantic invariance of our VAE compared to traditional VAEs using the SC metric. The calculation of the SC metric is shown in Algorithm~\ref{alg:sc}, where a higher SC value indicates better semantic consistency in the latents. Table~\ref{tab: Semantic consistency} demonstrate that our VAE can extract the similar semantics between two different variants of the same image.

\begin{table}[ht]
\vskip -0.1in
\centering
\setlength{\tabcolsep}{5pt}
\small % \scriptsize % \scriptsize %\footnotesize % \tiny
\caption{\textbf{Semantic Consistency.} A higher SC indicates that more semantic information is retained in the latent.}
% \begin{center}
% \begin{small}
% \begin{sc}
% \resizebox{7cm}{!}{
\begin{tabular}{l|cccccc}
\toprule
Data Aug. & Crop          & Flip          & GaussianBlur  & Grayscale     & All            \\ \midrule
SD-VAE    & 0.33          & 0.34          & 0.41          & 0.37          & 0.29           \\
Ours      & \textbf{0.45} & \textbf{0.44} & \textbf{0.46} & \textbf{0.47} &  \textbf{0.41} \\
\bottomrule
\end{tabular}
% }
% \end{sc}
% \end{small}
% \end{center}
\label{tab: Semantic consistency}
% \vskip -0.1in
\end{table}

The first experiment demonstrates that our VAE exhibits better semantic similarity between samples. The second experiment shows that our VAE has stronger feature attention within individual samples. The third experiment qualitatively indicates that our VAE achieves better semantic consistency with different data augmentations.

\subsection{Enhanced Generation Capability}

We compare the baseline models of DiT and SiT under the same training configuration, with Table~\ref{tab:FID} presenting the experimental results without using classifier-free guidance (cfg). The results indicate that under the same model parameters and training steps, diffusion models trained on ReaLS achieve significant performance improvements. Our approach requires no modifications to the diffusion model training process or additional network structures, providing a cost-free enhancement to the diffusion baseline, with an average FID improvement exceeding \textbf{15\%}.

\begin{table}[ht]
\caption{\textbf{FID Comparisons with Vanilla DiTs and SiTs.} Generate on ImageNet $256\times256$ without classifier-free guidance.}
\centering
\small % \scriptsize % \scriptsize %\footnotesize % \tiny
\resizebox{8cm}{!}{
\begin{tabular}{lcccccccc}
\toprule
Model & \multicolumn{1}{c}{VAE} & Params & \multicolumn{1}{c}{Steps} & \multicolumn{1}{c}{FID$\downarrow$} & \multicolumn{1}{c}{sFID$\downarrow$} & \multicolumn{1}{c}{IS$\uparrow$} & \multicolumn{1}{c}{Pre.$\uparrow$} & \multicolumn{1}{c}{Rec.$\uparrow$} \\ \midrule
DiT-B-2           & SD-VAE                  & 130M   & 400K                      & 43.5                    &      -                   & -                      & -                        & -                        \\
\textbf{DiT-B/2}  & \textbf{Ours}           & 130M   & \textbf{400K}             & \textbf{35.27}          & \textbf{6.30}            & \textbf{37.80}         & \textbf{0.56}            & \textbf{0.62}            \\ \midrule
SiT-B-2           & SD-VAE                  & 130M   & 400K                      & 33.0                    &     -                    &  -                     &  -                       & -                        \\
\textbf{SiT-B/2}  & \textbf{Ours}           & 130M   & \textbf{400K}             & \textbf{27.53}          & \textbf{5.49}            & \textbf{49.70}         & \textbf{0.59}            & \textbf{0.61}            \\
\textbf{SiT-B/2}  & \textbf{Ours}           & 130M   & \textbf{1M}               & \textbf{21.18}          & \textbf{5.42}            & \textbf{64.72}         & \textbf{0.63}            & \textbf{0.62}            \\
\textbf{SiT-B/2}  & \textbf{Ours}           & 130M   & \textbf{4M}               & \textbf{15.83}        & \textbf{5.25}             & \textbf{83.34}              & \textbf{0.65}             & \textbf{0.63}                         \\ \midrule
SiT-L-2           & SD-VAE                  & 458M   & 400K                      & 18.8                    &      -                   & -                      & -                        &  -                       \\
\textbf{SiT-L/2}  & \textbf{Ours}           & 458M   & \textbf{400K}             & \textbf{16.39}          & \textbf{4.77}            & \textbf{76.67}         & \textbf{0.66}                & \textbf{0.61}                \\ \midrule
SiT-XL-2          & SD-VAE                  & 675M   & 400K                      & 17.2                    &     -                    & -                      & -                        & -                        \\
\textbf{SiT-XL/2} & \textbf{Ours}           & 675M   & \textbf{400K}             & \textbf{14.24}          & \textbf{4.71}            & \textbf{83.83}         & \textbf{0.68}            & \textbf{0.62}            \\
\textbf{SiT-XL/2} & \textbf{Ours}           & 675M   & \textbf{2M}               &    \textbf{8.80}                     & \textbf{4.75}                         & \textbf{118.51}                       & \textbf{0.70}                         & \textbf{0.65}                        \\ \bottomrule
\end{tabular}
}
\label{tab:FID}
\vspace{-1em}
\end{table}

Table~\ref{tab: FID with cfg} displays the generation results of our model with cfg. In the comparative experiments with DiT-B/2 (80 epochs, cfg=1.5) and SiT-B/2 (200 epochs, cfg=1.5), the models trained on ReaLS consistently outperformed traditional VAE space, achieving better FID scores. In the SiT-XL/2 experiment, our model reached an impressive FID of \textbf{1.82} after a relatively low number of training epochs (i.e., 400 epochs).

% \arrayrulecolor{black!30}\midrule

\begin{table}[t]
\caption{\textbf{Generation on ImageNet $256\times256$ with classifier-free guidance.} $*[a,b]$ indicates the use of cfg with the guidance interval~\cite{kynkaanniemi2024applying}.}
\centering
\renewcommand{\arraystretch}{0.8} %0.8
\setlength{\tabcolsep}{1.6pt}
\tiny % \scriptsize % \scriptsize %\footnotesize % \tiny
\resizebox{8cm}{!}{
\begin{tabular}{lcccccc}
\toprule
Model & Epochs & FID$\downarrow$ & sFID$\downarrow$ & IS$\uparrow$ & Pre.$\uparrow$ & Rec.$\uparrow$ \\ \midrule

\multicolumn{7}{c}{\emph{GAN-based Generative Model}} \\ \arrayrulecolor{black!30}\midrule

BigGAN-deep~\cite{brock2018large}     & -       & 6.95            & 7.36             & 171.4        & 0.87                & 0.28             \\ 
StyleGAN-XL~\cite{stylegan}                & -       & 2.30            & 4.02             & 265.12       & 0.78                & 0.53             \\ \arrayrulecolor{black!100}\midrule

\multicolumn{7}{c}{\emph{Autoregressive Generative Model}} \\ \arrayrulecolor{black!30}\midrule
Mask-GIT~\cite{chang2022maskgit}    & 555    & 6.18            & -                & 182.1        & -                   & -                \\
MagViT-v2~\cite{yu2023language}   & 1080   & 1.78            & -                & 319.4        & -                   & -                \\
LlamaGen~\cite{sun2024llamagen}     & 300    & 2.18            & 5.97             & 263.3        & 0.81                & 0.58             \\
VAR~\cite{tian2024var}     & 350    & 1.80            & -                & 365.4        & 0.83                & 0.57             \\
MAR~\cite{li2024autoregressive-mar}     & 800    & 1.55            & -                & 303.7        & 0.81                & 0.62             \\ 

\arrayrulecolor{black!100}\midrule
\multicolumn{7}{c}{\emph{Diffusion Model}} \\ \arrayrulecolor{black!30}\midrule

ADM~\cite{beats_gan}    & -       & 10.94           & 6.02             & 100.98       & 0.69                & 0.63             \\
% ADM-U                      & -       & 7.49            & 5.13             & 127.49       & 0.72                & 0.63             \\
% ADM-G                      & -       & 4.59            & 5.25             & 186.70       & 0.82                & 0.52             \\
ADM-G, ADM-U               & 400     & 3.94            & 6.14             & 215.84       & 0.83                & 0.53             \\ \midrule
Simple Diff~\cite{hoogeboom2023simple}    & -    & 3.76            & -                & 171.6        & -                   & -                \\
Simple Diff(U-ViT, L) & 800       & 2.77            & -                & 211.8        & -                   & -                \\  \midrule

CDM~\cite{ho2022cascaded}                        & 2160   & 4.88            & -                & 158.71       & -                   & -                \\ 

U-ViT-H/2~\cite{uvit}                  & 240    & 2.29            & 5.68             & 263.9        & 0.82                & 0.57             \\ 
VDM++~\cite{kingma2024understanding}                      & 560    & 2.12            & -                & 267.7        & -                   & -                \\

\arrayrulecolor{black!100}\midrule
\multicolumn{7}{c}{\emph{Latent Diffusion Model}} \\ \arrayrulecolor{black!30}\midrule
LDM-8~\cite{rombach2022ldm}   & -       & 15.51           & -                & 79.03        & 0.65                & 0.63             \\
LDM-8-G                    & -       & 7.76            & -                & 209.52       & 0.84                & 0.35             \\
LDM-4                      & -    & 10.56           & -                & 103.49       & 0.71                & 0.62             \\
% LDM-4-G (cfg=1.25)         & -       & 3.95            & -                & 178.22       & 0.81                & 0.55             \\
LDM-4-G (cfg=1.50)         & 200       & 3.60            & -                & 247.67       & 0.87                & 0.48             \\ \midrule
RIN~\cite{jabri2022scalable}                        & -       & 3.42            & -                & 182.0        & -                   & -                \\ 

 \midrule

DiT-B/2 (cfg=1.5)~\cite{dit}          & 80     & 22.21           & -                & -            & -                   & -                \\
\rowcolor{blue!15} \textbf{DIT-B/2 + ReaLS (cfg=1.5)}   & \textbf{80}     & \textbf{19.44}           & \textbf{5.45}             & \textbf{70.37}        & \textbf{0.68}                & \textbf{0.55}             \\ 
DiT-XL/2                   & 1400   & 9.62            & 6.85             & 121.50       & 0.67                & 0.67             \\
DiT-XL/2 (cfg=1.25)      & 1400       & 3.22            & 5.28             & 201.77       & 0.76                & 0.62             \\
DiT-XL/2 (cfg=1.50)      & 1400       & 2.27            & 4.60             & 278.24       & 0.83                & 0.57             \\ 
SD-DiT~\cite{zhu2024sd}                     & 480    & 3.23            & -                & -            & -                   & -                \\ 
FasterDiT~\cite{yao2024fasterdit}                  & 400    & 2.03            & 4.63             & 264.0        & 0.81                & 0.60             \\ 
% 下面这三个结果，投稿版本一定要注释掉
FiT-XL/2~\cite{Lu2024FiT} & 400  & 4.21 & 10.01 & 254.87 & 0.84 & 0.51\\
FiTv2-XL~\cite{wang2024fitv2} & 400 & 2.26 & 4.53 & 260.95 & 0.81 & 0.59 \\
DoD-XL~\cite{yue2024dod} & 400 & 1.73 & 5.14 & 304.31 & 0.79 & 0.64 \\
\midrule

MaskDiT~\cite{zheng2023fast}                    & 1600   & 2.28            & 5.67             & 276.6        & 0.80                & 0.61             \\ 
MDT~\cite{gao2023masked}                        & 1300   & 1.79            & 4.57             & 283.0        & 0.81                & 0.61             \\
MDTv2                      & 1080   & 1.58            & 4.52             & 314.7        & 0.79                & 0.65             \\ \midrule

SiT-B/2 (cfg=1.5)~\cite{sit}          & 200      & 9.3           & -                & -            & -                   & -                \\
\rowcolor{blue!15} \textbf{SiT-B/2 + ReaLS (cfg=1.5)}   & \textbf{200}    & \textbf{8.39}            & \textbf{4.64}             & \textbf{131.97}       & \textbf{0.77}                & \textbf{0.53}             \\
% \rowcolor{blue!15} SiT-B/2 + ReaLS (cfg=2.0)*[0,0.75]  & 200    & 4.58            & 4.70             & 184.51       & 0.80                & 0.55             \\ 
\rowcolor{blue!15} SiT-B/2 + ReaLS (cfg=2.0)  & 650    & 4.38            & 4.52             & 239.08       & 0.86                & 0.46             \\ 
\rowcolor{blue!15} SiT-B/2 + ReaLS (cfg=2.0)*[0,0.75]  & 650    & 2.99            & 4.63             & 222.79       & 0.81                & 0.56             \\ 
\rowcolor{blue!15} SiT-B/2 + ReaLS (cfg=2.25)*[0,0.75]  & 650    & 2.74            & 4.58             & 251.02       & 0.83                & 0.54             \\ 
SiT-XL/2(cfg=1.5, ODE)     & 1400   & 2.15            & 4.60             & 258.09       & 0.81                & 0.60             \\
SiT-XL/2(cfg=1.5, SDE)     & 1400       & 2.06            & 4.49             & 277.50       & 0.83                & 0.59             \\ 
% \rowcolor{blue!15} SiT-XL/2 + ReaLS (cfg=2.0)*[0,0.75]  & 310    & 2.01            & 4.36             & 289.63       & 0.83                & 0.58             \\ 
% \rowcolor{blue!15} SiT-XL/2 + ReaLS (cfg=2.25)*[0,0.75]  & 310    & 2.50            & 4.38             & 319.66       & 0.84                & 0.56             \\ 
% \rowcolor{blue!15} SiT-XL/2 + ReaLS (cfg=1.25)  & 360    & 4.30            & 4.36             & 173.25       & 0.78                & 0.61             \\ 
% \rowcolor{blue!15} SiT-XL/2 + ReaLS (cfg=1.5)  & 360    & 2.93            & 4.25             & 226.31       & 0.82                & 0.56             \\ 
% \rowcolor{blue!15} SiT-XL/2 + ReaLS (cfg=1.75)  & 360    & 3.26            & 4.33             & 271.23       & 0.86                & 0.52             \\ 
% \rowcolor{blue!15} SiT-XL/2 + ReaLS (cfg=2.0)  & 360    & 4.31            & 4.53             & 306.87       & 0.89                & 0.47             \\ 
\rowcolor{blue!15} SiT-XL/2 + ReaLS (cfg=1.25)  & 400    & 4.18            & 4.39             & 175.16       & 0.77                & 0.60             \\ 
\rowcolor{blue!15} SiT-XL/2 + ReaLS (cfg=1.4)  & 400    & 3.08            & 4.29             & 208.60       & 0.81                & 0.58             \\ 
\rowcolor{blue!15} SiT-XL/2 + ReaLS (cfg=1.5)  & 400    & 2.83            & 4.26             & 229.59       & 0.82                & 0.56             \\ 
\rowcolor{blue!15} SiT-XL/2 + ReaLS (cfg=1.7)  & 400    & 3.02            & 4.31             & 266.70       & 0.86                & 0.52             \\ 
\rowcolor{blue!15} SiT-XL/2 + ReaLS (cfg=1.8)*[0,0.75]  & 400    & \cellcolor{yellow!20} 1.82            & 4.45             & 268.54       & 0.81                & 0.60             \\ 
\rowcolor{blue!15} SiT-XL/2 + ReaLS (cfg=2.0)*[0,0.75]  & 400    & 1.98            & 4.36             & 294.52       & 0.82                & 0.59             \\ 

\midrule

DiffiT*~\cite{hatamizadeh2025diffit}                    & -       & 1.73            & -                & 276.5        & 0.80                & 0.62             \\ \midrule
REPA~\cite{yu2024representation}           & 200    & 2.06            & 4.50             & 270.3        & 0.82                & 0.59             \\
REPA                       & 800    & 1.80            & 4.50             & 284.0        & 0.81                & 0.61             \\
REPA*                       & 800    & 1.42            & 4.70             & 305.7        & 0.80                & 0.65             \\
\bottomrule
\end{tabular}
}
\vspace{-2em}
\label{tab: FID with cfg}
\end{table}

\begin{figure*}[t]
    \centering
    \includegraphics[width=0.99\linewidth]{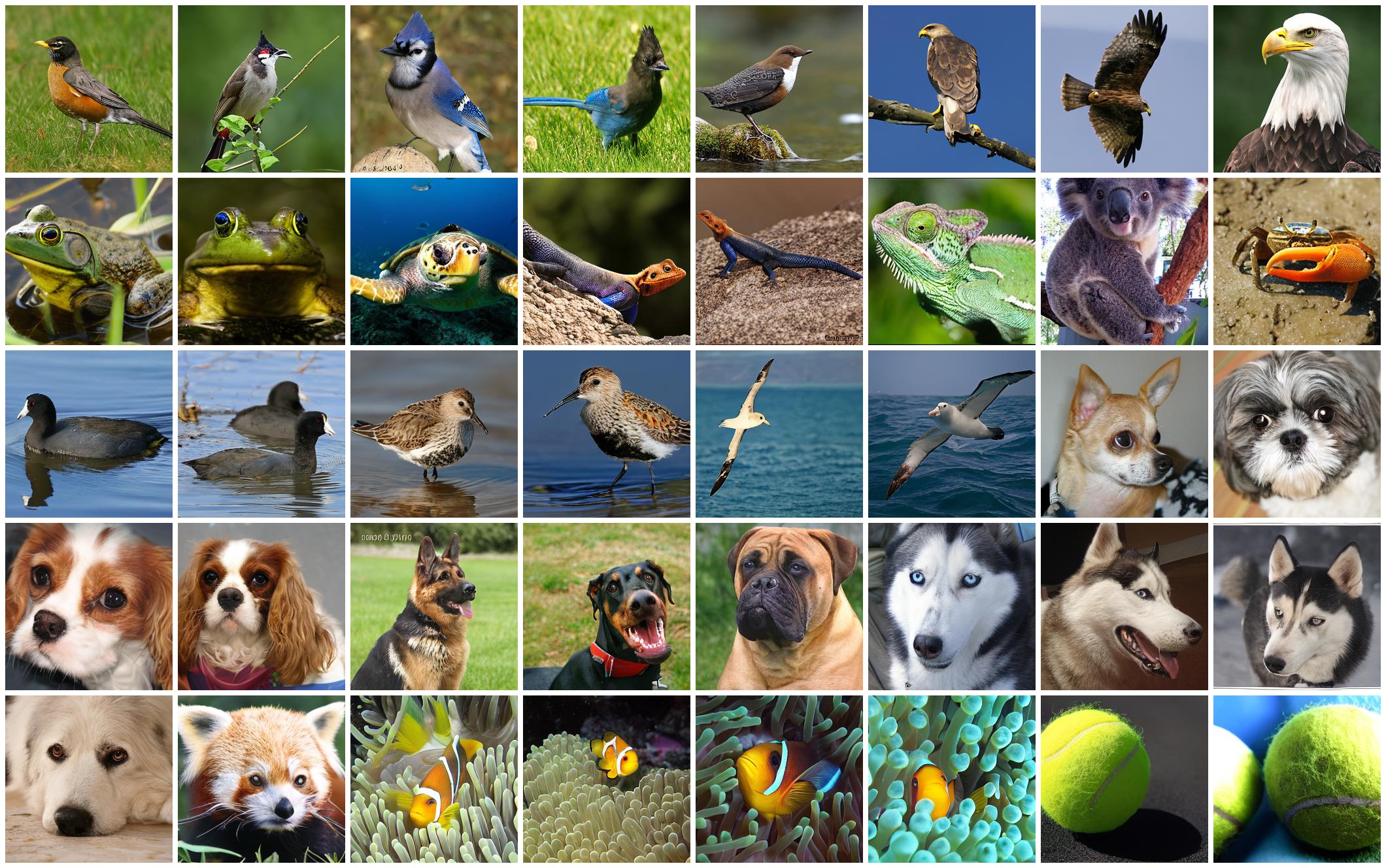}
    \vspace{-1em}
    \caption{\textbf{Visualization results} on ImageNet 256×256, from the SiT-XL/2 + ReaLS, with cfg=4.0.}
    \label{fig:visual_1}
    \vspace{-0.5cm}
\end{figure*}

\subsection{Downstream Tasks.}
\begin{figure*}
    \centering
    \includegraphics[width=0.99\linewidth]{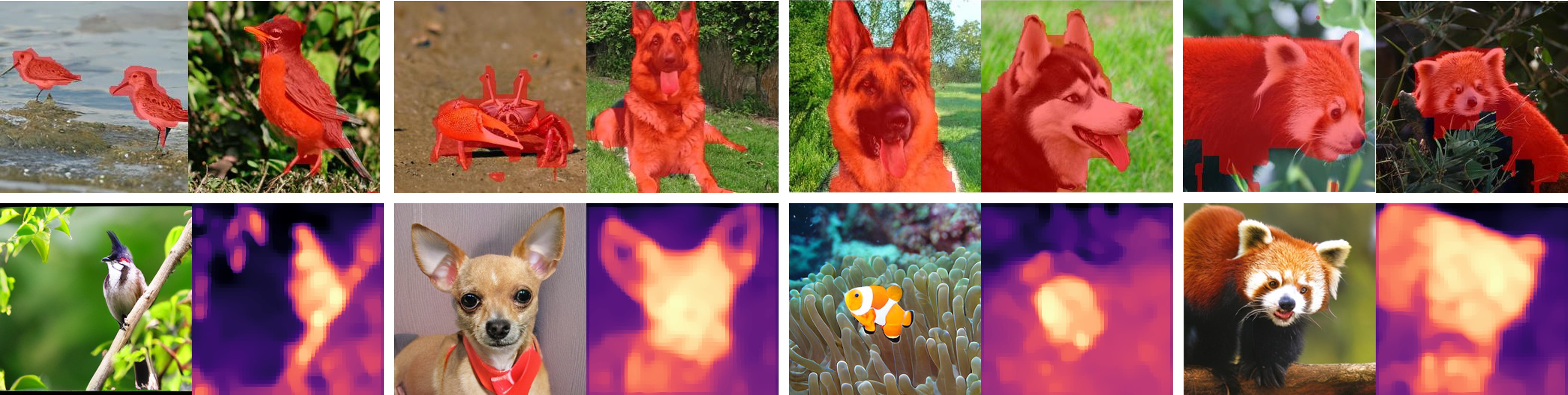}
    \vspace{-1em}
    \caption{\textbf{Training-free Downstream Tasks on Latents.} The diffusion model trained in the representation-aligned latent space naturally possesses stronger semantics, enabling more downstream tasks on latents. The latents generated by diffusion can obtain semantic features through the alignment network used during VAE training, and then multiple modalities of output can be achieved through the corresponding task heads. The first row displays the segmentation results, while the second row shows the depth estimation results.}
    \label{fig:visual_2}
\end{figure*}

By inputting the latents generated by the LDM model into the alignment network during VAE training, we obtain high-dimensional features with rich semantics similar to those of DINOv2. Then, through the segmentation head implemented in the \href{https://github.com/itsprakhar/Downstream-Dinov2}{Github repository}, we can achieve training-free generation of object masks.

Similarly, we use the depth estimation head of the MoGe~\cite{wang2024moge} to achieve training-free depth estimation for generated images. Figure~\ref{fig:visual_2} shows the segmentation mask and depth estimation generated when we use the SiT-XL/2 model to generate images. Downstream tasks involving perception in the latent space are still to be explored, and our approach presents a new possibility for unifying generation and perception.

\subsection{Ablation Studies}
\label{sec:kl analysis}
In the ablation study section, we aim to validate the impact of four key settings on the final generation quality. First, we investigate the effect of different KL weights in the latent loss discussed in Section~\ref{sec:loss}. Second, we explore whether aligning with DINOv2's patch features and cls features can each enhance the generation quality. Third, we examine whether the generation results align with different DINO models affect the final generation quality. Finally, we analyze the impact of different depths of the alignment network on the final generation quality.

\begin{figure}[ht]
    \centering
    \includegraphics[width=0.8\linewidth]{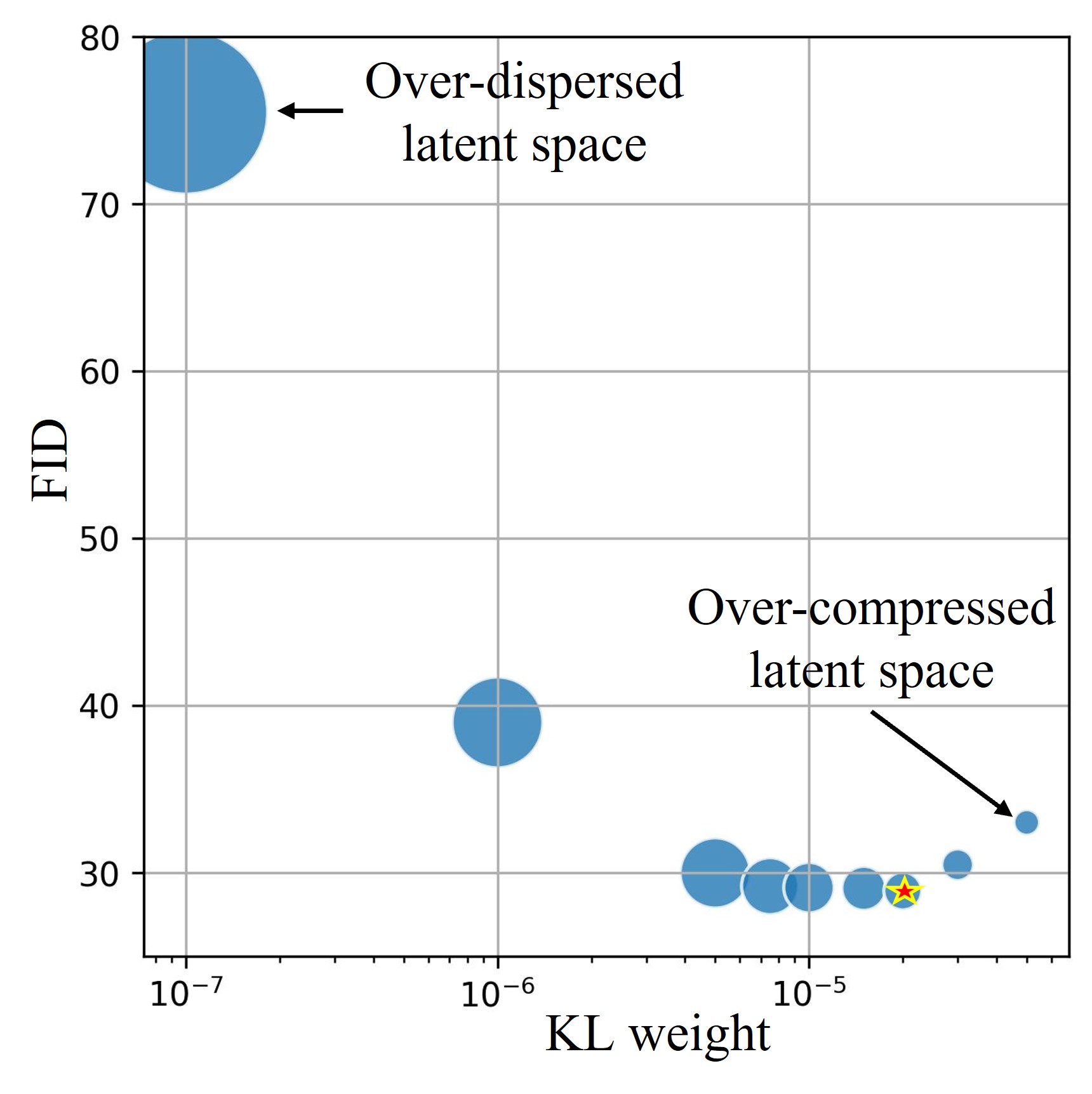}
    \vspace{-1em}
    \caption{\textbf{Impact of KL Constraint on Latent Space and FID.} As the KL weight increases from low to high, the FID initially decreases and then begins to rise again. The size of the point represents the standard deviation of the latent space.}
    \label{fig:kl}
    \vspace{-1.0cm}
\end{figure}

% kl weight增大，FID先降低，后上升
During model training, we found that the KL loss weight in the VAE significantly affects the final generation quality. Figure~\ref{fig:kl} illustrates the relationship between the KL weight and the FID of SiT-B/2 at 400k optimization steps. In Section~\ref{sec:loss}, we have analyzed how the KL loss constrains the integrity of the latent space, requiring the overall distribution to approximate a standard normal distribution. In contrast, the alignment loss constrains the position of each sample in this latent space, ensuring that samples with similar semantics are closer together. If we rely solely on alignment loss ($\lambda_k=0$ in the Equation~\ref{equ:latent loss}), the latent space becomes overly dispersed (large standard deviation), hindering generation. Conversely, a high KL weight imposes excessive constraints on the standard normal distribution (small standard deviation), limiting the alignment loss's effectiveness in semantic alignment. Therefore, we ultimately chose a KL weight of $\lambda_k=2e-5$ as our experimental setting.

% 使用dino的patch feature和cls feature对于生成都有益
Second, both DINOv2 features positively contribute to the final generation quality, as shown in Table~\ref{Ablation 1}. DINO's patch features represent local semantic characteristics of the image, while DINO's cls features reflect the overall characteristics. They guide the enhancement of semantic quality in the latent space at two different levels.

\begin{table}[ht]
\vskip -0.1in
\caption{\textbf{Impact of Aligning Different Features on Generation (400k Steps).}}
\centering
\small % \scriptsize % \scriptsize %\footnotesize % \tiny
% \begin{center}
% \begin{small}
% \begin{sc}
% \resizebox{6.5cm}{!}{
\begin{tabular}{l|ccccc}
\toprule
                                   & FID$\downarrow$ &  sFID$\downarrow$ & IS$\uparrow$ & Pre.$\uparrow$ & Rec.$\uparrow$ \\ \midrule
SiT-B-2                            & 33.0            & -                   & -            & -                   & -                \\
\  +DINO patch              & 28.91           & 5.65                & 48.46        & 0.59                & \textbf{0.62}             \\
\  \  +DINO cls & \textbf{27.53}           & \textbf{5.49}                & \textbf{49.70}        & \textbf{0.59}                & 0.61             \\ \bottomrule
\end{tabular}
% }
% % \end{sc}
% \end{small}
% \end{center}
\label{Ablation 1}
\vskip -0.1in
\end{table}

% 使用dino large的效果比dino base的效果好
Third, the generation results from DINOv2-large outperform those from DINOv2-base, as shown in Table~\ref{Ablation 2}. This is expected, as DINOv2-large features higher dimensionality and achieves better self-supervised learning metrics, resulting in richer semantic content and improved generation.

\begin{table}[ht]
\centering
\small
\vskip -0.1in
\caption{\textbf{Impact of Aligning Different DINO Models on Generation (400k Steps).}}
% \begin{center}
% \begin{small}
% \begin{sc}
% \resizebox{7cm}{!}{
\begin{tabular}{l|ccccc}
\toprule
  & FID$\downarrow$ & sFID$\downarrow$ & IS$\uparrow$ & Pre.$\uparrow$ & Rec.$\uparrow$ \\ \midrule
\multicolumn{1}{l|}{DINOv2-base} & 29.23           & 5.73             & 48.80        & 0.57                & \textbf{0.62}             \\
\multicolumn{1}{l|}{DINOv2-large} & \textbf{27.53}           & \textbf{5.49}             & \textbf{49.70}        & \textbf{0.59}                & 0.61             \\ 
\bottomrule
\end{tabular}
% }
% \end{sc}
% \end{small}
% \end{center}
% \vskip -0.1in
\label{Ablation 2}
\vspace{-1em}
\end{table}

% 使用两层的mlp做对齐的效果最好。
Finally, the alignment network composed of two linear layers outperforms both single-layer and four-layer configurations, as shown in Table~\ref{Ablation 3}. A shallow network results in poor alignment, while increasing the number of layers enhances the alignment network's nonlinear fitting ability, which may lead to overfitting of semantic information and consequently reduce the semantic content of the VAE's latent space. Therefore, opting for two linear layers as the alignment network is the optimal choice.

\begin{table}[ht]
\centering
\small % \scriptsize % \scriptsize %\footnotesize % \tiny
\vskip -0.1in
\caption{\textbf{Impact of Depth of Align Networks on Generation (400k Steps).}}
% \begin{center}
% \begin{small}
% % \begin{sc}
% \resizebox{7cm}{!}{
\begin{tabular}{l|ccccc}
\toprule
        & FID$\downarrow$ & sFID$\downarrow$ & IS$\uparrow$ & Pre.$\uparrow$ & Rec.$\uparrow$ \\ \midrule
1-layer & 33.66           & 7.12             & 42.96        & 0.53                & \textbf{0.64}    \\
2-layer & \textbf{27.53}  & \textbf{5.49}    & \textbf{49.70} & \textbf{0.59}     & 0.61             \\
4-layer & 29.00           & 6.25             & 47.83        & 0.58                & 0.62             \\ \bottomrule
\end{tabular}
% }
% % \end{sc}
% \end{small}
% \end{center}
\label{Ablation 3}
\vskip -0.1in
\end{table}

\subsection{Discussion}
\label{discussion}
\textbf{Better reconstruction does not necessarily lead to better generation.} Table~\ref{rec} shows the reconstruction metrics of our VAE on ImageNet $256\times256$. Although the reconstruction metrics of our VAE show a slight decline compared to SD-VAE, it provides a semantically rich latent space for the diffusion model, enhancing generation performance. This indicates that higher reconstruction quality does not necessarily lead to better generation results.

\begin{table}[ht]
\vskip -0.15in
\caption{\textbf{Reconstruction Metrics of VAEs.}}
\begin{center}
\begin{small}
% \begin{sc}
\resizebox{7.5cm}{!}{
\begin{tabular}{l|ccccc}
\toprule
Model   & rFID$\downarrow$ & PSNR$\uparrow$ & SSIM$\uparrow$ \\ \midrule
SD-VAE~\cite{rombach2022high}  & 0.74             & 25.68          & 0.820          \\
VQGAN~\cite{esser2021taming}   & 1.19             & 23.38          & 0.762          \\
Ours    & 0.85             & 23.45          & 0.768          \\
\bottomrule
\end{tabular}
}
% \end{sc}
\end{small}
\end{center}
\label{rec}
\vskip -0.1in
\end{table}

\textbf{The representation alignment in latent space and feature space can promote each other.} This paper focuses on aligning the VAE with image semantic representations to provide a better latent space with semantic priors for LDM. Additionally, some works have attempted to enhance image generation quality by incorporating semantics at the feature level, such as REPA~\cite{yu2024representation}. This work improves image generation quality by aligning the features of the diffusion model with image semantic representations.

Both this paper and REPA enhance generation quality through semantic augmentation; however, our approach emphasizes enhancing the latent space, while REPA enhances the LDM. This motivate us to explore the combination of both methods, investigating whether enhancing semantics in both the latent space and diffusion model features could further improve image generation quality.

Therefore, we trained the REPA model on ReaLS. The experimental results are as Table~\ref{repa}. It can be seen that the combination of the two methods yields better generation results than either method alone and significantly surpasses the baseline. After training for 1000k steps, the combined approach achieved a \textbf{30\%} improvement in FID compared to the baseline. These experimental findings further validate the importance of semantic alignment for generative tasks.

\begin{table}[ht]
\centering
\setlength{\tabcolsep}{3.6pt}
\small % \scriptsize % \scriptsize %\footnotesize % \tiny
\vskip -0.15in
\caption{\textbf{Combining ReaLS with REPA.} Representation alignment in both latent space and feature space enhances generation quality, with their combination yielding even better results.}
% \begin{center}
% \begin{small}
% % \begin{sc}
% \resizebox{7.5cm}{!}{
\begin{tabular}{lcccccc}
\toprule
       & Steps & FID$\downarrow$ & sFID$\downarrow$ & IS$\uparrow$ & Pre.$\uparrow$ & Rec.$\uparrow$ \\ \midrule
SiT-B-2 & 400k                & 33.0                    &     -                    &  -                     &  -                       & -                        \\
\ +REPA & 400k & 24.4  & -    & - & -     & -             \\
\ +ReaLS & 400k & 27.53          & 5.49            & 49.70         & 0.59            & 0.61           \\
\ +ReaLS, REPA & 400k & \textbf{23.40}          & \textbf{5.49}            & \textbf{57.55}         & \textbf{0.61}            & \textbf{0.62}           \\ \midrule
SiT-B-2 & 1000k                & 27.31                    &     -                    &  -                     &  -                       & -                        \\
\ +ReaLS, REPA & 1000k & \textbf{18.96}          & \textbf{5.54}            & \textbf{70.57}         & \textbf{0.64}            & \textbf{0.63} \\

\bottomrule
\end{tabular}
% }
% % \end{sc}
% \end{small}
% \end{center}
\label{repa}
% \vskip -0.1in
\end{table}

%% file: sections/conclusion.tex
\section{Conclusion}

The ability of generative models to produce high-quality content relies on effectively modeling the real world. A common type of generative model, the latent diffusion model, first encodes real-world samples into a latent space using a variational autoencoder (VAE), then learns the distribution of samples within that latent space. This generative paradigm implies that the modeling capability of the VAE directly influences the final generation results. Traditional VAEs compress images through reconstruction tasks, which only consider pixel-level local information and fail to capture the semantic priors of images effectively.

This paper enhances the semantic information in the latent space by aligning the VAE's latent space with semantic representation models. Experimental analysis shows that the latent space aligned with semantic representations exhibits better structural properties, characterized by increased diversity among different samples and enhanced correlations within the same sample. Generation experiments demonstrate that a semantically rich latent space is crucial for improving the generation quality of diffusion models. Furthermore, due to its rich semantics, diffusion models trained in this latent space inherently possess capabilities for various training-free perceptual downstream tasks.

%% file: sections/impact_statements.tex
% \section*{Impact Statements}

% This paper presents work whose goal is to advance the field of Machine Learning. There are many potential societal consequences of our work, none which we feel must be specifically highlighted here.

%% file: sections/supply.tex
\newpage
\appendix
\onecolumn
\section*{Appendix}

\subsection*{Loss Hyperparameters}

The complete form of the loss function is shown in Equation~\ref{loss all}. During actual training, we set $\lambda_g=0.1$, $\lambda_p=1.0$, $\lambda_{k}=2e-5$, $\lambda_{a}=1.0$.

\begin{equation}
\gL = \gL_{\text{MSE}} + \lambda_g \gL_{\text{GAN}} + \lambda_p \gL_{\text{perceptual}} + \lambda_{k} \gL_{\text{KL}} + \lambda_{a} \gL_{\text{align}}
\label{loss all}
\end{equation}

% \subsection*{Reconstruction}
% Table~\ref{rec} shows the reconstruction metrics of our VAE on ImageNet $256\times256$. Although the reconstruction metrics of our VAE show a slight decline compared to SD-VAE, it provides a semantically rich latent space for the diffusion model, enhancing generation performance. This indicates that higher reconstruction quality does not necessarily lead to better generation results.

% \begin{table}[ht]
% \vskip 0.15in
% \caption{\textbf{Reconstruction Metrics of VAEs.}}
% \begin{center}
% \begin{small}
% % \begin{sc}
% \resizebox{7.5cm}{!}{
% \begin{tabular}{l|ccccc}
% \toprule
% Model   & rFID$\downarrow$ & PSNR$\uparrow$ & SSIM$\uparrow$ \\ \midrule
% SD-VAE~\cite{rombach2022high}  & 0.74             & 25.68          & 0.820          \\
% VQGAN~\cite{esser2021taming}   & 1.19             & 23.38          & 0.762          \\
% Ours    & 0.85             & 23.45          & 0.768          \\
% \bottomrule
% \end{tabular}
% }
% % \end{sc}
% \end{small}
% \end{center}
% \label{rec}
% \vskip -0.1in
% \end{table}

\subsection*{Latent Normalization}

As illustrated in Figure~\ref{fig:kl}, the KL weight has a significant impact on the final generation results. Additionally, the KL weight plays a crucial role in the distribution of the latent variables, as different KL weights can significantly affect the mean and variance of the latent space, as shown in Table~\ref{mean std}. Therefore, during the training of the diffusion model, to maintain consistency with the training of SD-VAE, we normalized the latents to the same numerical range as that of SD-VAE.

\begin{table}[ht]
\vskip 0.15in
\caption{\textbf{The distribution of latent space changes with kl weight.} Calculate the value by sampling 10,000 samples from the ImageNet $256\times256$, encoded by VAE aligned with DINOv2-base.}
\begin{center}
\begin{small}
% \begin{sc}
\resizebox{7.5cm}{!}{
\begin{tabular}{c|ccccc}
\toprule
kl weight   & mean & std & min & max \\ \midrule
SD-VAE   & 0.29287 & 4.58407 & -65.730 & 68.3175 \\
0        & 1.32886 & 5.42394 & -48.074 & 38.0941 \\
1.00E-06 & -0.0463 & 1.3918  & -8.8677 & 7.2072  \\
5.00E-06 & 0.00251 & 1.0678  & -7.9659 & 9.8775  \\
7.50E-06 & -0.0059 & 1.03842 & -8.2779 & 11.661  \\
1.00E-05 & -0.0092 & 1.02894 & -7.0984 & 9.31086 \\
1.50E-05 & -0.0091 & 1.02723 & -9.1763 & 11.1787 \\
2.00E-05 & 0.00394 & 1.0266  & -15.916 & 17.6631 \\
3.00E-05 & -0.0148 & 1.0256  & -15.192 & 16.1069 \\
5.00E-05 & -0.0002 & 1.00952 & -12.118 & 13.2441 \\
\bottomrule
\end{tabular}
}
% \end{sc}
\end{small}
\end{center}
\label{mean std}
\vskip -0.1in
\end{table}

In terms of normalization methods, we experimented with $\text{std}$ normalization and $\text{max}-\text{min}$ normalization, as presented in Table~\ref{norm}. The experiments indicate that using $\text{max}-\text{min}$ normalization yields better generation performance.

\begin{table}[ht]
\vskip 0.15in
\caption{\textbf{The impact of different normalization methods of latents on the quality of generation.} Use kl weight=5e-6, SiT-B/2 model at 400k optimization steps.}
\begin{center}
\begin{small}
% \begin{sc}
\resizebox{4cm}{!}{
\begin{tabular}{c|ccccc}
\toprule
normalization method & FID \\ \midrule
$\text{std}$ & 40 \\
$\text{max}-\text{min}$  & 32 \\
\bottomrule
\end{tabular}
}
% \end{sc}
\end{small}
\end{center}
\label{norm}
\vskip -0.1in
\end{table}

Specifically, during encoding, the latents are scaled by Equation~\ref{encode}, and during decoding, the latents generated by diffusion are scaled by Equation~\ref{decode}.

\begin{equation}
\left\{
\begin{matrix}
z=(z-\text{mean}_\text{ours})/(\text{max}_\text{ours}- 
\text{min}_\text{ours}) \\
z=z\times(\text{max}_\text{SD-VAE}-\text{min}_\text{SD-VAE})+\text{mean}_\text{SD-VAE} 
\end{matrix}
\right.
\label{encode}
\end{equation}

\begin{equation}
\left\{
\begin{matrix}
z=(z-\text{mean}_\text{SD-VAE})/(\text{max}_\text{SD-VAE}- 
\text{min}_\text{SD-VAE}) \\
z=z\times(\text{max}_\text{ours}-\text{min}_\text{ours})+\text{mean}_\text{ours} 
\end{matrix}
\right.
\label{decode}
\end{equation}

where $\text{mean}_\text{ours}=-0.016722$, $\text{max}_\text{ours}=10.762420$, $\text{min}_\text{ours}=-6.862830$, $\text{mean}_\text{SD-VAE}= 0.292873$, $\text{max}_\text{SD-VAE}=68.317589$, $\text{min}_\text{SD-VAE}=-65.730583$ for VAE aligned with DINOv2-large-reg (kl weight=2e-5).